\pdfoutput=1

\documentclass[11pt]{article}

\usepackage[table,xcdraw]{xcolor}
\usepackage[final]{acl}

\usepackage{times}
\usepackage{latexsym}

\usepackage[T1]{fontenc}

\usepackage[utf8]{inputenc}

\usepackage{microtype}

\usepackage{inconsolata}

\usepackage{graphicx}
\usepackage{amsmath}
\usepackage{makecell}
\usepackage{enumitem}

\usepackage[utf8]{inputenc} 
\usepackage[T1]{fontenc}    
\usepackage{hyperref}       
\usepackage{url}            
\usepackage{booktabs}       
\usepackage{amsfonts}       
\usepackage{nicefrac}       
\usepackage{microtype}      
\usepackage{xurl}
\title{GraphCheck: Breaking Long-Term Text Barriers with \\Extracted Knowledge Graph-Powered Fact-Checking}

\author{
 \textbf{Yingjian Chen\textsuperscript{1}\thanks{Equal contributions}},
 \textbf{Haoran Liu\textsuperscript{2}\footnotemark[1]\,},
 \textbf{Yinhong Liu\textsuperscript{3}},
 \textbf{Jinxiang Xie\textsuperscript{1}},
 \textbf{Rui Yang\textsuperscript{4}},
 \\
 \textbf{Han Yuan\textsuperscript{4}},
 \textbf{Yanran Fu\textsuperscript{1}},
 \textbf{Peng Yuan Zhou\textsuperscript{5}},
 \textbf{Qingyu Chen\textsuperscript{6}},
 \\
 \textbf{James Caverlee\textsuperscript{2}},
 \textbf{Irene Li\textsuperscript{1}}\thanks{Corresponding author},
\\
\\
 \textsuperscript{1}University of Tokyo,
 \textsuperscript{2}Texas A\&M University,
 \textsuperscript{3}University of Cambridge,
 \\
 \textsuperscript{4}Duke-NUS Medical School,
 \textsuperscript{5}Aarhus University,
 \textsuperscript{6}Yale University.
\\
 \small{
   \textbf{} \href{mailto:irene.li@weblab.t.u-tokyo.ac.jp}{irene.li@weblab.t.u-tokyo.ac.jp}
 }
}

\begin{document}
\maketitle

\begin{abstract}
Large language models (LLMs) are widely used, but they often generate subtle factual errors, especially in long-form text. These errors are fatal in some specialized domains such as medicine. 
Existing fact-checking with grounding documents methods face two main challenges: (1) they struggle to understand complex multihop relations in long documents, often overlooking subtle factual errors; (2) most specialized methods rely on pairwise comparisons, requiring multiple model calls, leading to high resource and computational costs. 
To address these challenges, we propose \textbf{\textit{GraphCheck}}, a fact-checking framework that uses extracted knowledge graphs to enhance text representation. 
Graph Neural Networks further process these graphs as a soft prompt, enabling LLMs to incorporate structured knowledge more effectively. 
Enhanced with graph-based reasoning, GraphCheck captures multihop reasoning chains that are often overlooked by existing methods, enabling precise and efficient fact-checking in a single inference call.
Experimental results on seven benchmarks spanning both general and medical domains demonstrate up to a 7.1\% overall improvement over baseline models.
Notably, GraphCheck outperforms existing specialized fact-checkers and achieves comparable performance with state-of-the-art LLMs, such as DeepSeek-V3 and OpenAI-o1, with significantly fewer parameters.\footnote{Our code is available at \url{https://github.com/Yingjian-Chen/GraphCheck}.}
\end{abstract}

\section{Introduction}

\begin{figure}[t!]
    \centering
    \includegraphics[width=0.48\textwidth]{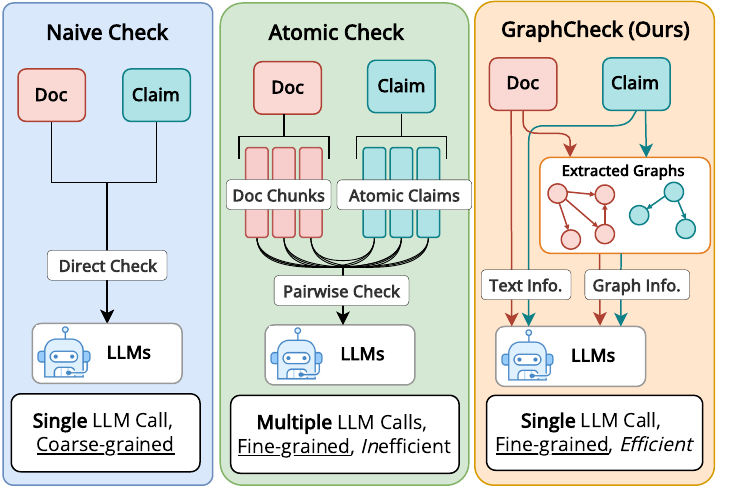}
    \caption{Comparison of fact-checking methods. \textbf{Naive Check} performs a single-pass evaluation but often misses detailed factual errors. \textbf{Atomic Check} ensures fine-grained verification by checking atomic facts individually but is inefficient due to multiple LLM calls. In contrast, our \textbf{GraphCheck} achieves fine-grained fact-checking in a single call, significantly improving efficiency while maintaining accuracy.}
    \label{fig:overview}
    \vspace{-2mm}
\end{figure}

Large language models (LLMs)~\cite{hurst2024gpt, dubey2024llama}, have demonstrated powerful generative capabilities in various domains~\cite{yang2023large,Lee2024ASO,liu2023recprompt,yang2024ascle}. 
However, due to limitations in training data and the lack of integration of domain-specific knowledge, LLMs often ``\textit{hallucinate}'' factual errors or inaccurate information~\cite{mckenna-etal-2023-sources,zhang2023language,Gao2023EvaluatingLL}. As LLMs prioritize linguistic fluency and contextual relevance in their generation processes, the generated content may appear convincing while lacking factuality~\cite{huang2023survey, ramprasad2024analyzing, yang-etal-2024-kg}. 
This issue is particularly concerning in specialized domains like medicine, where factual errors can compromise patient safety, leading to misdiagnoses, inappropriate treatments, and, in severe cases, life-threatening consequences \cite{Ahsan2023RetrievingEF,yang-etal-2024-kg}. 
Therefore, ensuring the reliability and factual accuracy of LLM outputs is essential~\citep{liu2024aligning}.

We consider the task of fact-checking claims against grounding documents, where the goal is to assess factual consistency based on provided textual evidence \cite{tang2024minicheck}.
Given the high cost and time demands of manual verification, modern fact-checking methods have shifted to automated approaches using LLMs or natural language inference (NLI) models \cite{fu2023gptscore, Kim2024CanLP}. 
Standard LLM-based checking methods take a straightforward approach by directly feeding documents and claims into LLM for fact-checking judgment (Figure~\ref{fig:overview}, left).
However, when dealing with long-form documents, they often struggle to capture complex entity relations and overlook subtle inconsistencies given large volumes of information.
Additionally, long prompts may exceed the LLM's context window, causing potential loss of relevant details and limiting the model from effective fact-checking.
To address this, specialized methods \cite{zha2023alignscore, min2023factscore, liu2024evaluating} decompose long documents into smaller chunks and break claims into atomic facts, enabling fine-grained evaluation at the price of computational cost and efficiency (Figure~\ref{fig:overview}, middle).

To address the problem of long text fact-checking, we propose \textit{\textbf{GraphCheck}} (Figure~\ref{fig:overview}, right), a graph-enhanced framework using extracted knowledge graphs (KGs) to capture multi-hop logical relations between entities, enhancing both global coherence and fine-grained understanding in long texts.
We employ Graph Neural Networks (GNNs)~\cite{velivckovic2017graph, yun2019graph} to encode these graph structures and integrate the graph embeddings into LLM inputs~\cite{he2024g, tian2024graph, jin2024large}. 
The direct comparison between the extracted document and claim graphs enables fine-grained factual verification in an LLM inference.
The GNNs are trained on our curated general-domain synthetic graph data based on MiniCheck~\cite{tang2024minicheck} training set, while the LLMs remain frozen. 
Empirically, we find that despite being trained on general-domain data, our model achieves improved performance not only on general-domain datasets but also on medical-domain datasets, demonstrating that its graph-enhanced reasoning ability generalizes across domains.
We also provide this dataset as a benchmark for future research, allowing the training and evaluation of graph-based fact-checking.

In summary, our contributions are:

\begin{itemize}[noitemsep, topsep=1pt]
    \itemsep 0em
    \item \textbf{Pioneering Graph Reasoning for LLM Fact-Checking.} We propose GraphCheck, the first graph reasoning-enhanced LLM framework for fact-checking with grounding documents, ensuring fine-grained factual accuracy with high efficiency. 
    \item \textbf{Enabling Fine-grained Explainability.}
    Our method enhances explainability by identifying the key entity relationships the model focuses on during fact-checking, ensuring a clear and verifiable reasoning process.
    \item \textbf{Providing a Benchmark for Graph-based Fact-Checking Models.} 
    We introduce a synthetic dataset that pairs text with its corresponding extracted knowledge graph, enabling the training and evaluation of KG-enhanced fact-checking models.
    \item \textbf{Empirical Findings.}
    We demonstrate the effectiveness and efficiency of GraphCheck, achieving a 7.1\% improvement over the base model in fact-checking across extensive general and medical benchmarks.
\end{itemize}

\section{Related Work and Background}

\textbf{Methods in Detecting Hallucination.} 
Recent fact-checking research \cite{yuan2023zero, kim2023factkg} uses Retrieval-Augmented Generation (RAG) \cite{Fan2024ASO,yang2025retrieval} and external knowledge bases like DBpedia \cite{lehmann2015dbpedia} and Wikidata \cite{vrandevcic2014wikidata} to verify generated claims by retrieving structured or semi-structured data.

Another line of research \cite{manakul2023selfcheckgpt, mundler2023self} focuses on verifying factual consistency using LLMs with grounding documents. These approaches harness LLMs’ reasoning and language capabilities to fact-check claims against textual evidence. While effective for short texts, they often fail to capture fine-grained inconsistencies in longer documents, limiting their accuracy.
Our work builds on this second setting, aiming to improve fact-checking performance on long texts by enhancing LLMs with structured graph-based reasoning.


\begin{figure*}[t!]
    \centering
    \includegraphics[width=1\textwidth]{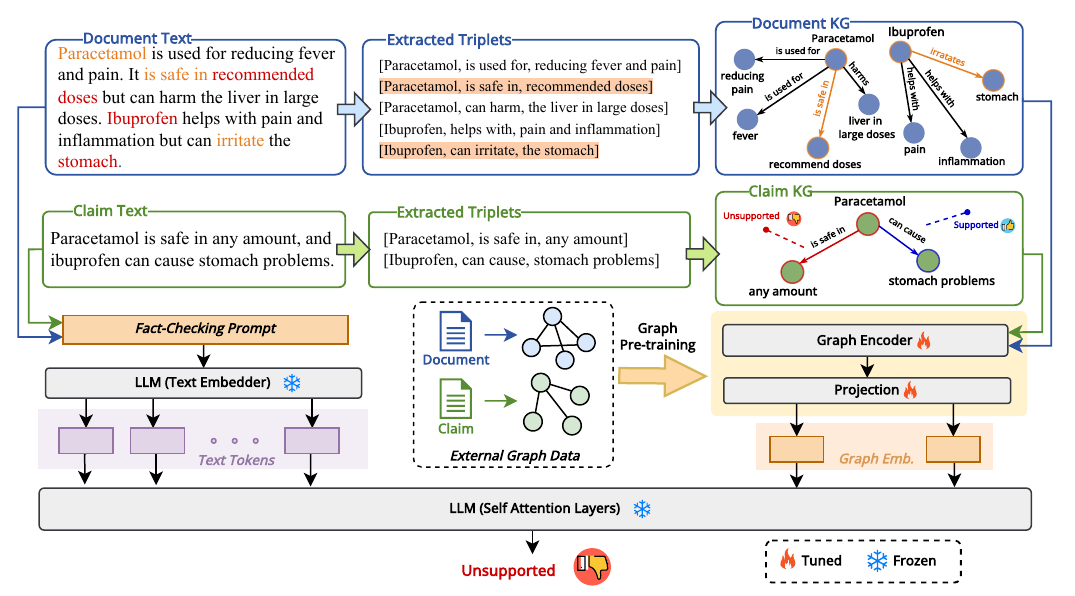}
    \caption{An illustration of the GraphCheck framework. Firstly, an LLM extracts entity-relation triples from both the claim and the document to construct KGs, respectively. 
    A GNN pre-trained with external text graph data is then used to obtain graph embeddings from both KGs. 
    These graph embeddings, combined with the text embeddings, are fed into an LLM for final fact-checking. 
    This approach enables the LLM to perform fine-grained fact-checking by leveraging key triples in the KG (highlighted) alongside the text information.}
    \label{fig:pipline}
\end{figure*}

\noindent\textbf{Fact-Checking on Long Texts.}
To address the challenge of capturing detailed errors in long texts, recent methods have shifted towards using fine-grained units for fact-checking. 
Methods like FactScore \cite{min2023factscore}, MiniCheck \cite{tang2024minicheck}, and ACUEval \cite{wan2024acueval} focus on extracting atomic units from the generated text to enable fine-grained fact verification. 
However, these fine-grained fact-checking methods often require multiple calls to verify each unit or triple, especially for long texts, which greatly increases computational cost and time.
In contrast, our approach uses KGs to model complex entity relationships in long texts, enabling fine-grained verification in a single call. This avoids repetitive calls and significantly improves efficiency.

\noindent\textbf{Graph-based Methods for Enhancing Factuality. }
Previous graph-based fact-checking methods have primarily focused on isolated triple evaluations or document-level encoding, often overlooking the global graph structure and topological information. 
GraphEval \cite{liu2024evaluating} extracts triples from claims and evaluates their factual consistency individually using a pretrained NLI model. 
However, it also relies on pairwise comparisons and does not incorporate the overall graph structure, limiting its ability to capture complex relationships. 
FactGraph \cite{ribeiro2022factgraph} employs graph encoders to process documents and summary semantic graphs extracted via OpenIE. It then combines text and graph embeddings through an MLP for the final prediction. 
However, as a pre-LLM method, it lacks the powerful contextual reasoning ability of modern models. 
AMRFact \cite{qiu_amrfact_2024} leverages AMR graphs to represent document structures and guide factual summarization generation, focusing on structured summarization rather than direct fact verification.
Unlike previous methods, our approach integrates a trainable GNN with an LLM~\cite{tian2024graph, he2024g}, combining long-form contextual understanding with structured knowledge from extracted KGs. By incorporating graph reasoning, our model captures complex entity relationships and logical structures, enabling fine-grained fact verification in a single comparison. 
This enhanced reasoning ability allows the model to generalize effectively to specialized domains.
An extended discussion on how LLMs understand knowledge graphs is provided in Appendix~\ref{appendix:discuss}.


\section{GraphCheck}
In this section, we introduce our \textit{GraphCheck} framework, which is designed for efficient fact-checking. 
Intuitively, GraphCheck first extracts structural information from KGs to enrich the input text and then leverages an LLM for verification. 
GraphCheck contains three main steps: 
(1) Given a source document $D$ and a generated claim $C$, we extract knowledge triples from them and construct corresponding KGs. 
(2) A trainable GNN encodes the entire graph, generating comprehensive graph embeddings. (3) These embeddings, along with the document and claim texts, are fed into a verifier LLM, with frozen parameters, enabling single-call fine-grained fact-checking with the help of structured graph information, as shown in Figure \ref{fig:pipline}.

\subsection{Graph Construction}
\label{graph_construction}
To construct the KGs, we extract triples in the form of \(\{source, relation, target\}\) from the text, where each entity and relation captures key semantic information. To achieve this, an LLM is employed to automatically identify and extract these triples. The detailed prompt used for triple extraction is provided in Appendix~\ref{sec:prompts}. 
Building on the extracted triplets, we construct a directed graph \( G = (\mathcal{V}, \mathcal{E})\). 
Here, $\mathcal{V} = \{\mathbf{v}_i\}_{ i= 1, \dots, n}$ is the set of node (entity) features, where each $\mathbf{v}_i$ denotes the feature vector for node $i$.
$\mathcal{E} = \{\mathbf{e}_{ij}\}_{  i,j=1, \dots, n}$ is the set of edge (relation) features, where $\mathbf{e}_{ij}$ denotes the edge feature vector for an edge from node $i$ to node $j$.
The node features and edge features from textual attributes are encoded using Sentence-Transformers.\footnote{\url{https://huggingface.co/sentence-transformers/all-roberta-large-v1}}
For a given generated claim $C$ and its source document $D$ we extract the corresponding graphs \( G_{\text{C}} \) and \( G_{\text{D}} \). 


\subsection{GraphCheck Verification}
\noindent{\textbf{Graph Encoding.}}
We encode the extracted KGs with a GNN. Specifically, for the $l$-th GNN layer updates node features based on the message passing scheme as:
\begin{equation*}
    \textbf{v}^{l+1}_i = \text{UPDATE}\left( \textbf{v}^l_i,
    \sum_{j\in\mathcal{N}_i}\text{MESSAGE}\left(\textbf{v}^l_j, \textbf{e}_{ji}\right)\right),
    \label{eq:mp}
\end{equation*}
where $\mathcal{N}_i$ denotes the set of node $i$'s neighbors,
and UPDATE and MESSAGE functions are implemented by neural networks.
The final graph embeddings $\mathbf{h}_g$ are obtained with the GNN output layer, which is implemented with a READOUT function: 
\begin{equation*} 
    \textbf{h}_\text{g} = \text{READOUT}\left(\{\textbf{v}^L_i\}_{i=1,\dots,n}\right).
    \label{eq:readout}
\end{equation*}
Here, $\textbf{v}^L_i$ indicates the feature vector of node $i$ at the last layer. 
Specifically, the READOUT function includes a summation function to capture a global representation of the graph.

\noindent{\textbf{Text Encoding.}}
For a given generated claim $C$ and the source document $D$, we concatenate them following the verifying template shown in Appendix~\ref{sec:prompts}, and pass the rendered prompt into the verifier LLM to obtain the text embedding $\textbf{h}_t$.

\noindent\textbf{Graph Projection.}
To align the graph features with the verifier LLM's textual embedding space, we employ a projector module $P$. This module maps the extracted graph features of claim $\mathbf{h}_{\text{g}}^{\text{C}}$ and document $\mathbf{h}_{\text{g}}^{\text{D}}$ into the LLM's embedding space, resulting in the projected graph embeddings $\mathbf{\tilde{h}}_{\text{g}}^{\text{C}}$ and $\mathbf{\tilde{h}}_{\text{g}}^{\text{C}}$ for the claim and document, respectively.

\noindent{\textbf{Fact-Checking.}}
After obtaining the projected graph embeddings, \( \mathbf{\tilde{h}}_{\text{g}}^{\text{C}} \) and \( \mathbf{\tilde{h}}_{\text{g}}^{\text{D}} \), along with the text embedding \( \mathbf{h}_t  \), we concatenate them to construct the final input representation, which is then fed into the LLM self-attention layers for fact-checking:
\begin{equation*}
y = \text{LLM}(\mathbf{\tilde{h}}_{\text{g}}^{\text{C}}, \mathbf{\tilde{h}}_{\text{g}}^{\text{D}}, \mathbf{h}_t ),
\end{equation*}
where \( y \in \{\text{``support''}, \text{``unsupport''}\} \). The model consider both the structured and textual information to determine whether the document supports the claim.

By incorporating graph embeddings, our method effectively captures complex multi-hop logic relations in long text while ensuring efficient fact-checking. The knowledge graph, which encodes entity relationships within the entire text, assists the LLM in detecting factual inconsistencies that may be overlooked when relying solely on text.

\begin{table}[t]
    \centering
    \resizebox{0.48\textwidth}{!}{
    \begin{tabular}{lcccc}
        \toprule
        Dataset & Size & \( \text{Doc}_{len} \) & \( \text{Claim}_{len} \) & Neg\% \\
        \midrule
        \textit{\textbf{General Domain}} &&&& \\
        AggreFact-Xsum & 558  & 324 & 23 & 48.9\% \\
        AggreFact-CNN  & 558  & 500 & 55 & 10.2\% \\
        Summeval       & 1600 & 359 & 63 & 18.4\%\\
        ExpertQA       & 3702 & 432 & 26 & 19.8\%\\
        \midrule
        \textit{\textbf{Medical Domain}} &&&& \\
        COVID-Fact     & 4086 & 72  & 12 & 68.3\%\\
        PubHealth      & 1231 & 77  & 14 & 51.3\%\\
        SCIFact        & 809  & 249 & 12 & 58.9\%\\
        \bottomrule
    \end{tabular}}
    \caption{Statistics of Benchmark Datasets. We report the size of each benchmark, the average text length of source documents and generated claims, and the proportion of negative samples.}
    \label{tab:datasets}
\end{table}

\section{Experimental Setup}

\begin{table*}[t]
    \centering
    \resizebox{1.0\textwidth}{!}{
    \begin{tabular}{lccccccccccc|c}
    \toprule
         \multicolumn{1}{c}{} &  \multicolumn{4}{c}{\textbf{General Domain}} && \multicolumn{3}{c}{\textbf{Medical Domain}} & \multicolumn{1}{c}{} \\
         \cmidrule(lr){2-6} \cmidrule(lr){7-9}
         \textbf{Method} & \makecell{\textbf{AggreFact}\\\textbf{-Xsum}} & \makecell{\textbf{AggreFact}\\\textbf{-CNN}} & \textbf{Summeval} & \textbf{ExpertQA} && \textbf{COVID-Fact} & \textbf{SCIFact} & \textbf{PubHealth} &  \makecell{\textbf{Overall} \\Avg. (\%)} \\
        \midrule
         \textit{\textbf{Large-scale LLMs}}\textsuperscript{\textcolor{blue}{*}} &&&&&&&&& \\
         GPT-4 \cite{Achiam2023GPT4TR}     & 75.4 & 60.7 & 69.7 & 59.6 && 73.8 & 83.3 & 73.2 &  70.8\\
         GPT-4o \cite{hurst2024gpt}    & 76.4 & 66.8 & 76.3 & 58.3 && 62.6 & 83.2 & 67.0 &  70.1\\
         OpenAI o1 \cite{jaech2024openai}   & 74.8 & 65.3 & 70.5 & 58.8 && 75.9 & 90.3 & 74.8 & 72.9\\
         Claude 3.5-Sonnet \cite{claude35sonnet} & 75.7 & 68.8 & 77.3 & 58.8 && 73.8 & 87.2 & 73.8 & \underline{73.6}\\
         DeepSeek-V3 671B\cite{liu2024deepseek} & 74.6 & 63.2 & 68.3 & 58.5 &&  75.9 & 89.1 & 72.9 &  71.7\\
         \midrule \midrule
         \textit{\textbf{Small-scale LLMs}} &&&&&&&&& \\
         Llama3 8B \cite{dubey2024llama}  & 53.4 & 51.3 & 51.7 & 51.3 && 58.1 & 62.2 & 70.7 & 57.0\\
         Qwen2.5 7B \cite{yang2024qwen2} & 53.2 & 45.3 & 58.5 & 53.6 && 59.2 & 53.5 & 59.1 & 54.7\\
         Llama3.3 70B \cite{dubey2024llama} & 60.1 & 53.5 & 57.6 & 54.3 && 69.0 & \cellcolor[HTML]{FFFFB3}85.7 & \cellcolor[HTML]{FFB3B3}76.9 & 65.3\\
         Qwen2.5 72B \cite{yang2024qwen2} & 55.6 & 49.9 & 53.4 & 54.1 && \cellcolor[HTML]{FFD9B3}69.9 & 85.6 & \cellcolor[HTML]{FFD9B3}76.7 & 63.6\\
         \midrule
         \textit{\textbf{Specialized Fact-checking Methods}} &&&&&&&& \\
         AlignScore \cite{zha2023alignscore} & 68.0 & 54.1 & 62.2 & \cellcolor[HTML]{FFD9B3}59.3 && 66.5 & 71.7 & 64.4 & 63.7\\
         ACUEval \cite{wan2024acueval}    & 55.5 & 50.0 & 53.7 & \cellcolor[HTML]{FFFFB3}57.5 && 64.7 & 79.9 & 62.9 & 60.6\\
         MiniCheck \cite{tang2024minicheck}  & \cellcolor[HTML]{FFFFB3}70.8 & \cellcolor[HTML]{FFFFB3}63.7 & \cellcolor[HTML]{FFB3B3}74.8 & 57.4 && 65.9 & 78.1 & 66.3 & \cellcolor[HTML]{FFFFB3}68.1\\
         GraphEval \cite{sansford2024grapheval}  & 67.6 & \cellcolor[HTML]{FFB3B3}69.5 & \cellcolor[HTML]{FFFFB3}69.7 & 56.0 && 60.7 & 68.4 & 63.7 & 65.1\\
         \midrule
         \textit{\textbf{Ours}} &&&&&&&&& \\
         GraphCheck-Llama3.3 70B & \cellcolor[HTML]{FFB3B3}72.9 & 62.4 & 67.3 & \cellcolor[HTML]{FFB3B3}60.3 && \cellcolor[HTML]{FFB3B3}71.5 & \cellcolor[HTML]{FFB3B3}89.4 & \cellcolor[HTML]{FFFFB3}73.6 & \cellcolor[HTML]{FFB3B3}71.1\\
         GraphCheck-Qwen 72B & \cellcolor[HTML]{FFD9B3}72.1 & \cellcolor[HTML]{FFD9B3}66.5 & \cellcolor[HTML]{FFD9B3}71.0 & 57.2 && \cellcolor[HTML]{FFFFB3}69.7 & \cellcolor[HTML]{FFD9B3}86.4 &  71.7 & \cellcolor[HTML]{FFD9B3}70.7\\
    \bottomrule
    \end{tabular}}
    \caption{Balanced accuracy of fact-checkers across all benchmarks, covering both general and medical domains. Methods are categorized into \textit{\textbf{Large-scale LLMs}}\textsuperscript{\textcolor{blue}{*}} | \textit{\textbf{Small-scale LLMs}} | \textit{\textbf{Specialized Fact-checking Methods}} | \textit{\textbf{Ours}}. The \colorbox[HTML]{FFB3B3}{top-1}, \colorbox[HTML]{FFD9B3}{top-2}, and \colorbox[HTML]{FFFFB3}{top-3} performances for each dataset among models smaller than Large-scale LLMs are highlighted, while the best-performing results within Large-scale LLMs are \underline{underlined}.}
    \label{tab:main_result}
\end{table*}

\subsection{Datasets}
\label{sec:dataset}
\noindent\textbf{Training Dataset.} To train a GNN for extracting KG information, we use the \{claim, document, label\} pairs from MiniCheck dataset \cite{tang2024minicheck} with 14K synthetic samples. We use Claude-3.5-Sonnet~\cite{claude35sonnet} to extract KG triples for claims and documents, constructing graphs for each pair. The final training dataset is structured as \{\( C \), \( D \), \( G_{\text{C}} \), \( G_{\text{D}} \), label\}. 
The dataset is split into training, validation, and test sets in a 6:2:2 ratio for model training and evaluation.

\noindent\textbf{Evaluation Benchmarks.} 
Our work mainly focuses on fact-checking tasks that involve long-term text, as shown in Table \ref{tab:datasets}. Therefore, we adopt widely used datasets like AggreFact-CNN \cite{tang-etal-2023-understanding}, AggreFact-XSum \cite{tang-etal-2023-understanding}, and SummEval \cite{10.1162/tacl_a_00373}, all of which include lengthy documents. To assess our method's performance in open-domain scenarios, we also incorporate the long-text question-answering dataset ExpertQA \cite{malaviya2023expertqa}. Furthermore, we evaluate our method on medical datasets, including SciFact \cite{wadden2020fact}, COVID-Fact \cite{saakyan2021covid}, and PubHealth \cite{kotonya2020explainable}, which provide specialized medical domain information. More details are shown in Appendix \ref{appendix_benchmark}.

\subsection{Baselines}
To comprehensively evaluate our method, we compare it against various fact-checkers, categorized into large-scale LLMs, small-scale LLMs, and specialized fact-checking methods.

Large-scale LLMs\footnote{We consider Large-scale LLMs as models with more than 300 B parameters.} include GPT-4 \cite{Achiam2023GPT4TR}, GPT-4o \cite{hurst2024gpt}, OpenAI o1 \cite{jaech2024openai}, Claude 3.5-Sonnet \cite{claude35sonnet}, and the largest open source model DeepSeek-V3 671B \cite{liu2024deepseek}. For small-scale LLMs, we include Llama3 8B, Llama3.3 70B \cite{dubey2024llama}, Qwen2.5 7B, and Qwen2.5 72B \cite{yang2024qwen2}. For specialized fact-checking methods, we include AlignScore \cite{zha2023alignscore} and fine-grained fact-checkers like MiniCheck~\cite{tang2024minicheck} and ACUEval \cite{wan2024acueval}. Additionally, we also consider graph-based methods, namely GraphEval \cite{sansford2024grapheval} and GraphRAG \cite{edge2024local}.

\subsection{Evaluation Metric}
Considering the data imbalance in some benchmarks, models biased towards a particular class in predictions may not reflect their true performance. To address this, we follow previous approached~\cite{liu-etal-2023-towards-interpretable, tang-etal-2023-understanding} and calculate balanced accuracy (BAcc). For more implementation details, please refer to Appendix~\ref{implementation_details}.

\section{Results and Analysis}
\subsection{Main Results}
Table \ref{tab:main_result} presents the BAcc of our proposed method, GraphCheck, compared to LLMs and specialized fact-checkers across general and medical domain benchmarks. The results show that our proposed GraphCheck achieves strong performance, reaching an overall BAcc of 71.1\% across all benchmarks. Specifically, among large-scale LLMs, Claude 3.5-Sonnet achieves the best overall performance. Our method outperforms GPT-4 and GPT-4o and comes close to the most advanced large-scale models, including OpenAI o1, Claude 3.5-Sonnet, and the latest open-source model DeepSeek-V3 671B, while operating at a smaller scale and significantly lower cost. Interestingly, GPT-4o underperforms on the medical datasets COVID-Fact and PubHealth, which contain shorter texts, even scoring lower than GPT-4. For small-scale LLMs, our method achieves improvements of 5.8\% and 7.1\% over the similarly sized base models, Llama3.3 70B and Qwen2.5 72B, respectively. For Specialized Fact-checking Methods, GraphCheck outperforms all methods, achieving 10.5\%, 3\%, and 6\% improvements over ACUEval, MiniCheck, and GraphEval, respectively. Notably, compared to methods that require multiple calls, our method achieves superior performance with a single model call. 
Although GraphRAG is not typically used for fact-checking, its popularity motivated us to adapt it for this purpose. A detailed analysis of these adaptations is provided in the Appendix~\ref{graphrag}.

\begin{figure}[t]
    \centering
    \includegraphics[width=0.95\linewidth]{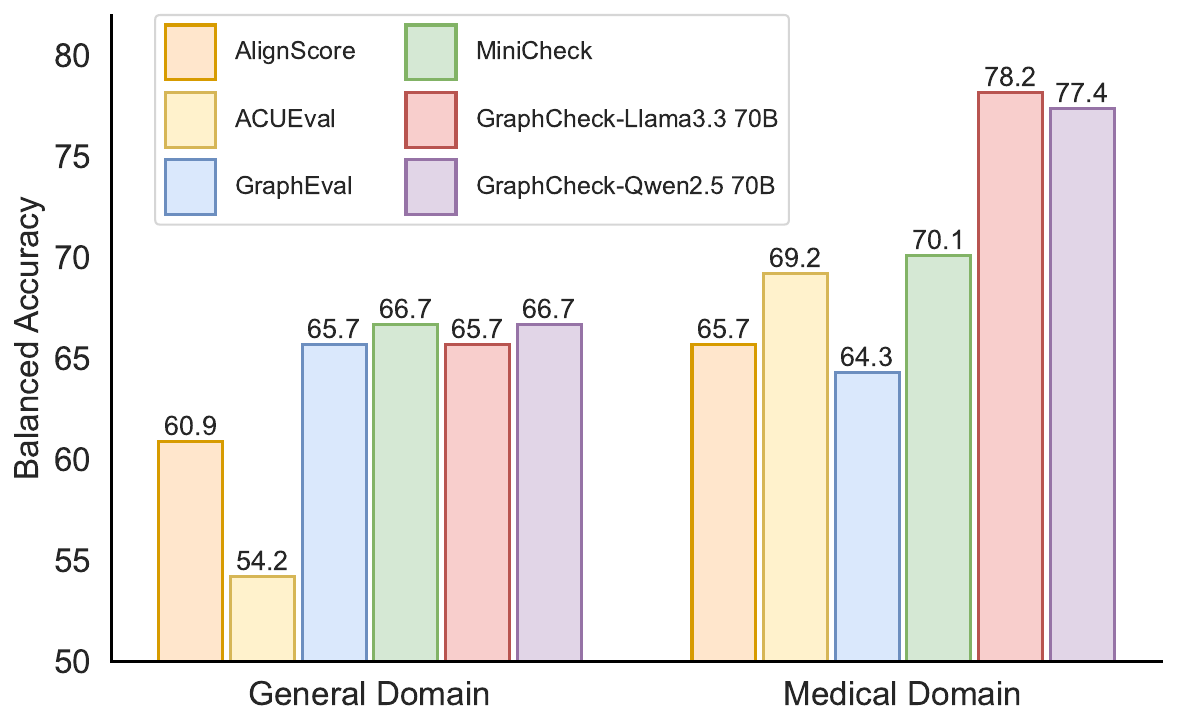}
    \caption{Average BAcc across general and medical domains. We compare our method with the specialized fact-checking methods in general domain (AggreFact-XSum, AggreFact-CNN, Summeval, ExpertQA) and medical domain (COVID-Fact, PubHealth, SCIFact).}
    \label{fig:barchart}
\end{figure}

In particular, our method achieves a BAcc of 60.3\% on ExpertQA, surpassing all models. This may be because GraphCheck can extract complex logical relations from graph data. However, our method underperforms on AggreFact-CNN and Summeval, which contain longer claims (average length > 50) and include more factual details. This makes knowledge triplets extraction more challenging, as some important information may be lost during the process, affecting subsequent fact-checking.

\noindent{\textbf{Performance Analysis in Different Domains.}}
To evaluate the effectiveness of our method across different domains, we compare it with other specialized fact-checking methods in both general and medical domains, as shown in Figure \ref{fig:barchart}. In the general domain, our method matches the performance of approaches like MiniCheck and GraphEval, which require multiple calls. However, in the medical domain, our method significantly outperforms these methods, achieving an 8.1\% improvement over Minicheck. This demonstrates the strong generalization ability of our method, when other methods perform limited in the medical domain, our method still maintains strong performance.

\begin{figure}[tbp]
    \centering
    \includegraphics[width=0.9\linewidth]{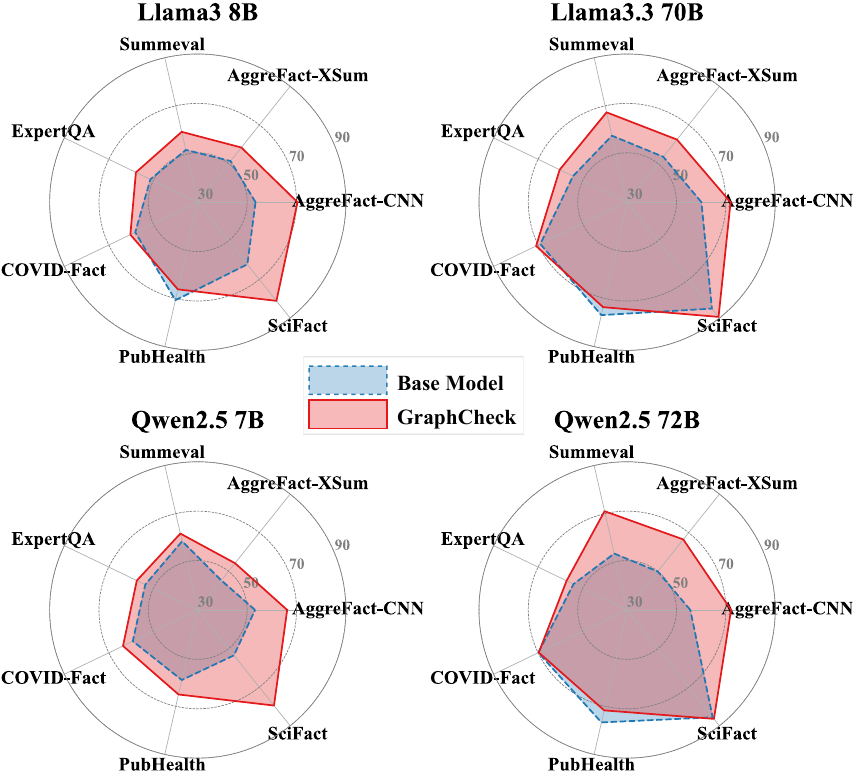}
    \caption{The BAcc of the base model and the proposed GraphCheck architecture across all seven benchmarks for Llama3 8B, Llama3.3 70B, Qwen2.5 7B, Qwen2.5 72B models. The blue-shaded region represents the base model performance, while the red-shaded region highlights the enhanced performance with GraphCheck.}
    \label{fig:ablation1}
\end{figure}

\begin{table*}[t]
    \centering
    \resizebox{1.0\textwidth}{!}{
    \begin{tabular}{lccccccccccc|c}
    \toprule
         \textbf{Method} & \makecell{\textbf{AggreFact}\\\textbf{-Xsum}} & \makecell{\textbf{AggreFact}\\\textbf{-CNN}} & \textbf{Summeval} & \textbf{ExpertQA} && \textbf{COVID-Fact} & \textbf{SCIFact} & \textbf{PubHealth} &  \makecell{\textbf{Overall} \\Avg. (\%)} \\
        \midrule
         Llama3.3 70B & 60.1 & 53.5 & 57.6 & 54.3 && 69.0 & 85.7 & 76.9 & 65.3\\
         Llama3.3 70B + KG Text & 60.6 & 53.6 & 57.8 & 54.4 && 71.2 & 88.8 & \textbf{78.5} & 66.4 \\
         GraphCheck-Llama3.3 70B + KG Text & 66.3 & 60.2 & 63.8 & 55.6 && 71.2 & 87.9 & 70.4 & 67.9 \\
         GraphCheck-Llama3.3 70B & \textbf{72.9} & \textbf{62.4} & \textbf{67.3} & \textbf{60.3} && \textbf{71.5} & \textbf{89.4} & 73.6 & \textbf{71.1} \\
    \bottomrule
    \end{tabular}}
    \caption{Balanced accuracy of the base model Llama3.3 70B and our proposed GraphCheck, with or without incorporating KG as prompt text, across all benchmarks. The best-performing results are \textbf{bold}.}
    \label{tab:kg_prompt}
\end{table*}

\subsection{Ablation Studies}
\noindent{\textbf{Impact of Additional Graph Information.}}
\label{ablation_graph}
To evaluate the effectiveness of incorporating graph information, we compare (1) the base LLM models with (2) our proposed GraphCheck, which is based on these models. As shown in Figure \ref{fig:ablation1}, our approach has a significant improvement on both lightweight models (Llama3 8B\footnote{\url{https://huggingface.co/meta-llama/Meta-Llama-3-8B-Instruct}}, Qwen2.5 7B\footnote{\url{https://huggingface.co/Qwen/Qwen2.5-7B-Instruct}}) and larger models (Llama3.3 70B, Qwen2.5 72B). Specifically, as shown in Table \ref{tab:datasets}, our method achieves significant improvement on relatively long-text datasets AggreFact-XSum, AggreFact-CNN, and Summeval. 

\begin{table}[b!]
    \centering
    \resizebox{0.48\textwidth}{!}{
    \begin{tabular}{lccccc}
        \toprule
         Method & \makecell{COVID\\-Fact} & \makecell{Pub\\Health} & \makecell{MiniCheck \\TestSet} & Reveal & \makecell{\textbf{Overall} \\Avg. (\%)} \\
        \midrule
        Qwen2.5 72B & 69.9 & 76.7 & 69.9 & 85.9 & 75.6 \\
        Llama3.3 70B & 69.0 & 76.9 & 71.0 & 86.8 & 75.9 \\
        AlignScore & 66.5 & 64.4 & 71.9 & 85.3 & 72.0 \\
        MiniCheck & 65.9 & 66.3 & $\text{{-}{-}}^{\textcolor{blue}{*}}$ & 88.8 & $\text{{-}{-}}^{\textcolor{blue}{*}}$ \\
        GraphEval & 60.7 & 63.7 & 72.1 & 89.8 & 71.6 \\
        \midrule
        \makecell[l]{GraphCheck-\\Qwen2.5 72B} & 69.7 & 73.6 & 78.4 & 88.0 & \textbf{77.4} \\
        \midrule
        \makecell[l]{GraphCheck-\\Llama3.3 70B} & 71.5 & 71.7 & 81.2 & 89.7 & \textbf{78.5} \\
        \bottomrule
    \end{tabular}}
    \caption{Additional evaluations across four short-text datasets: COVID-Fact, PubHealth, the MiniCheck test set, and Reveal~\cite{jacovi2024chain}. \textcolor{blue}{*} denotes that the MiniCheck model is trained on the full dataset, including the test set, so the results are omitted for fair comparison.}
    \label{tab:short_text}
\end{table}

A similar result is observed on the relatively longer SCIFact dataset, where our approach significantly enhances lightweight models. However, for larger models, which can already handle longer texts effectively, the improvement is much more limited. The above results demonstrate the effectiveness of our method, showing that GraphCheck enhances the ability of models to handle long-text fact-checking tasks. Additionally, graph information is essential for effectively capturing complex logical relations within the text.

\noindent{\textbf{Evaluating the Use of KGs as Prompt Text.}}
Given that extracted KGs can be directly used as the prompt text for LLMs, as employed in G-Retriever~\cite{he2024g}, we conduct ablation studies to assess the impact of using the knowledge graph as the prompt text for LLMs, as shown in Table~\ref{tab:kg_prompt}. Experimental results show that directly adding KGs text into the prompt yields only marginal improvement over the base model Llama3 70B and even results in performance degradation when integrated into our proposed GraphCheck. This is mainly because LLMs struggle to capture fine-grained factual errors directly from the long textual inputs, and the large extracted knowledge graph text may further exacerbate this issue. In contrast, our proposed GraphCheck only integrates KGs through a GNN, which enables more effective integration of structured graph information.

\begin{figure}[b!]
    \centering
    \includegraphics[width=0.80\linewidth]{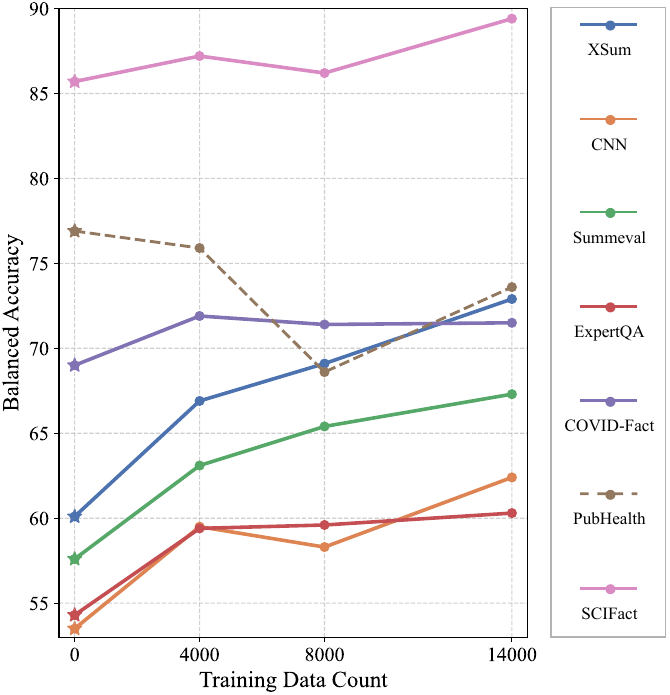}
    \caption{Balanced accuracy comparison across different training data sizes on all benchmarks. The baseline model performance is marked at 0 on the $x$-axis.}
    \label{fig:ablation2}
\end{figure}

\begin{figure*}[t!]
    \centering
    \includegraphics[width=1\linewidth]{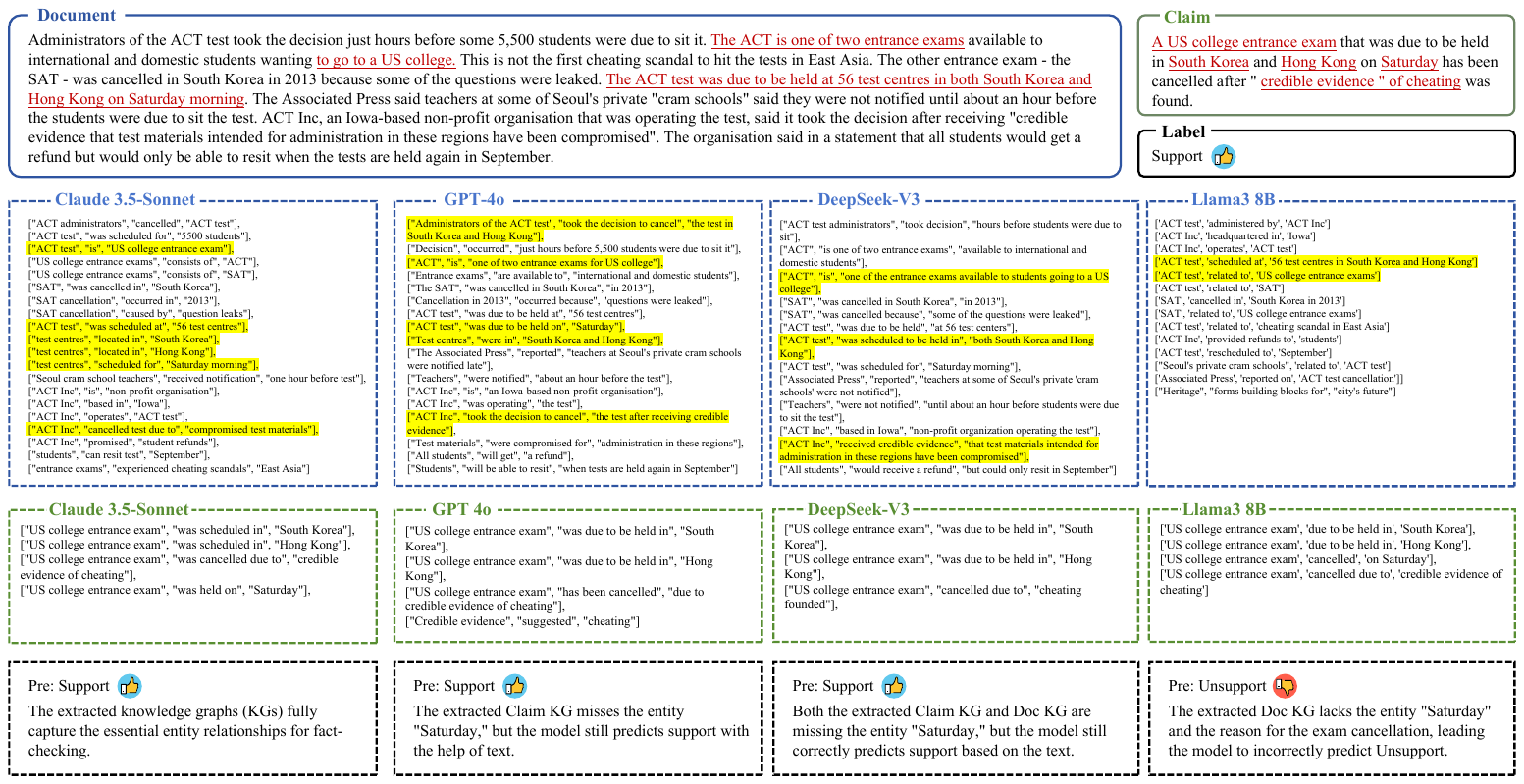}
    \caption{Example Analysis of the Impact of Knowledge Graph (KG) Quality on Model Prediction Results. The figure illustrates the influence of KGs extracted by four different models (Claude 3.5-Sonnet, GPT-4o, DeepSeek-V3, Llama 8B) on the performance of GraphCheck fact-checking.}
    \label{fig:kg_analysis}
\end{figure*}

\noindent{\textbf{Evaluation on Short-Text Fact-Checking.}}
Although GraphCheck is specifically designed for long-term text fact-checking, we also provide evaluations on short texts (length < 150). To do so, we conducted additional experiments on four short-text datasets, as shown in Table~\ref{tab:short_text}. The MiniCheck TestSet refers to the test split obtained from the 14K dataset using a 6:2:2 ratio (see Section~\ref{sec:dataset}). The results demonstrate that our proposed  GraphCheck maintains robust performance on short-text fact-checking tasks, outperforming existing specialized methods and base models, indicating its effectiveness across varying text lengths. Although the performance of  GraphCheck shows degradation only on the PubHealth dataset compared to the base models, this is primarily because Qwen2.5 72B and Llama-3.3 70B already achieve exceptionally high performance on this dataset, even surpassing OpenAI’s o1 (as shown in Table~2).

\noindent{\textbf{Impact of Training Data Sizes.}}
To evaluate the impact of training data size on the fact-checking performance of GraphCheck-Llama3.3 70B, we conducted experiments across all benchmarks, as shown in Figure \ref{fig:ablation2}. The results demonstrate a general upward trend in model performance as the amount of training data increases. Specifically, significant improvements can be observed on the long-text datasets AggreFact-XSum, AggreFact-CNN, Summeval, and SCIFact. Among them, XSum exhibits the largest improvement, increasing from 60.1\% to 72.9\%, while CNN and Summeval also achieve approximately 10\% improvements. In contrast, for the short-text datasets, our method shows only a slight improvement on COVID-Fact, while on PubHealth, performance gradually declines. These results further validate the conclusion drawn in Section \ref{ablation_graph}.

From the above results, we can observe that as the training data size increases, the overall model performance shows an upward trend. Therefore, we believe that further increasing the data size could continue to enhance the performance of our proposed GraphCheck. Additional experiments on different graph-building methods and GNN architectures are provided in Appendix~\ref{appendix: additional}.

\noindent{\textbf{Impact of Generated Knowledge Graph Quality.}} 
Due to the inherent randomness in extracting entity-relationship triples from text using LLMs, we conducted an experiment to assess how the quality of KGs generated from text impacts the model's final fact-checking results, as illustrated in Figure \ref{fig:kg_analysis}. For shorter generated claims, the triples extracted by the four models show minimal differences, except for occasional missing details by GPT-4o and DeepSeek-V3. These missing have minor effects on fact-checking results, as the models also relied on the original text for verification. In contrast, for longer document texts, there are significant differences in the quality of the triples generated by the models. Specifically, the triplets extracted by the Llama 8B model lacked crucial details, such as the time (\textit{"Saturday"}) and the reason for the exam cancellation. The loss of key information could potentially turn the KG into interference during fact-checking, ultimately leading to incorrect results. On the other hand, while the language expressions of the triples extracted by GPT-4, Claude 3.5, and DeepSeek-V3 are different, they all capture the essential details and still ensure that the fact-checker makes the right prediction.

These findings indicate that for short texts, the quality of the extracted KG has minimal impact on fact-checking performance, as models still rely on the original text for verification. However, for long-form documents, the completeness of the KG is critical. If the KG lacks key fact-checking information, it misleads the model rather than assists in verification. This is because longer texts make it more difficult for the model to extract essential details directly, increasing its dependence on the KG. In such cases, an incomplete or inaccurate KG introduces noise and ultimately compromises fact-checking accuracy. Conversely, if the missing information is irrelevant to the verification process, its absence does not affect the result.

\begin{figure}[t!]
    \centering
    \includegraphics[width=1\linewidth]{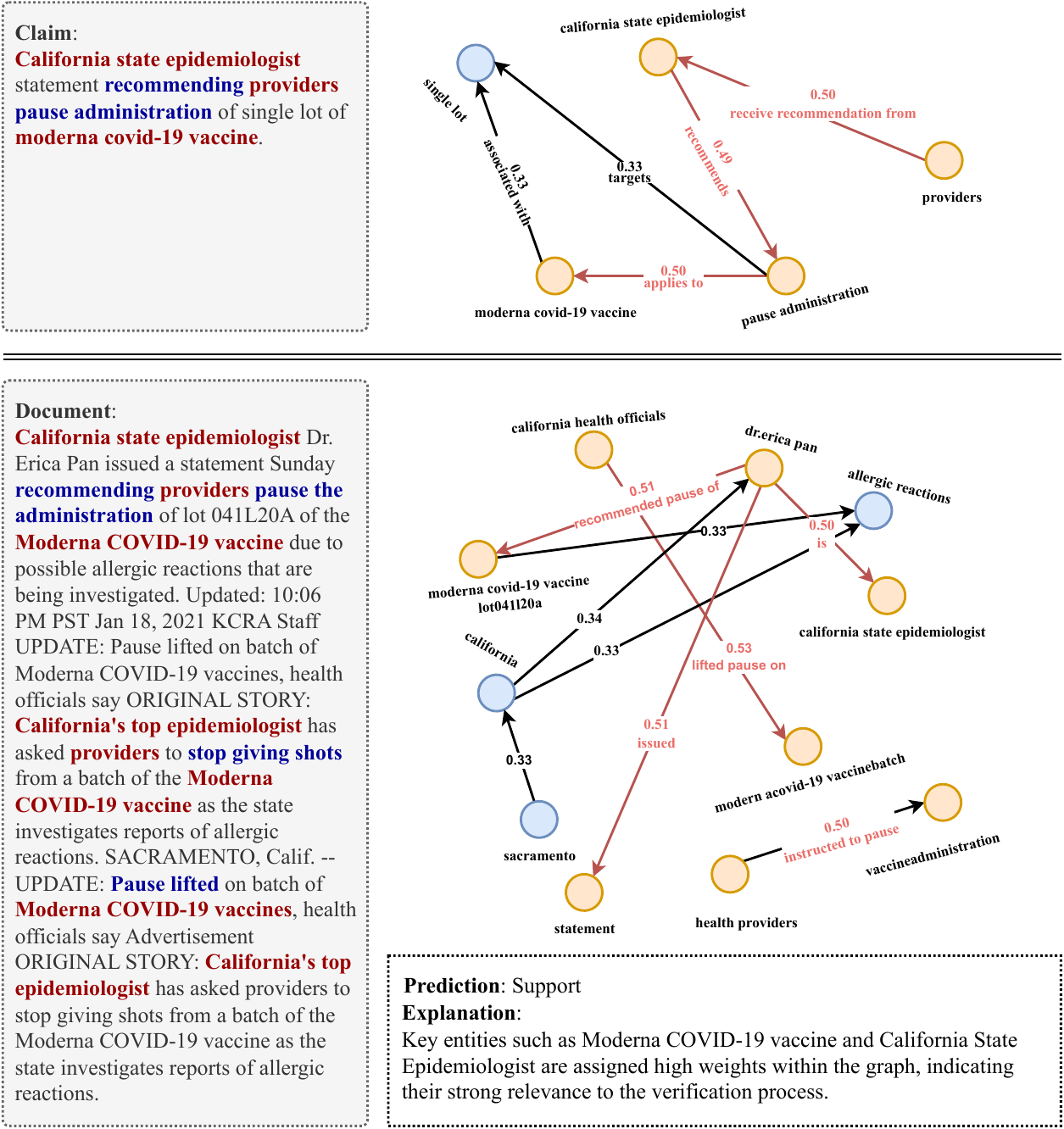}
    \caption{A case study in the medical domain. Connection weights in the KG are visualized to highlight key relationships primarily used by the model for fact-checking. Key entities and relationships in the text are marked in red and blue, while high-weighted nodes in the KG are highlighted in orange-yellow.}
    \label{fig:case_study}
\end{figure}

\subsection{Rethink Graph Importance on Long-Form Fact-checking}
We conduct a case study in the medical domain to demonstrate how our method uses KGs to help LLMs in the fact-checking process. We also showcase how our approach provides clear and interpretable explanations for the final checking results, as shown in Figure \ref{fig:case_study}. For each edge in the graph, we visualize its connection weight to show the attention distribution learned by the GNN model. The results indicate that the model selectively focuses on specific edges by assigning higher attention weights, emphasizing key relationships in the graph. Notably, these high-weight triplets correspond to key relations that align with the fact-checking requirements. For instance, the triplets (\textit{Dr. Erica Pan, is, California state epidemiologist}) and (\textit{Dr. Erica Pan, recommended pause of, Moderna COVID-19 vaccine}) in the document KG capture key information needed to verify the claim.

\noindent{\textbf{Explainability.}} 
This visualization not only highlights the key information the model relies on, but also improves the explainability of its fact-checking process. By revealing which relationships receive higher attention, it becomes easier to understand how the model makes its final decision and incorporates graph reasoning into its predictions. This explainability is particularly important in the medical domain, where fact-checking requires a clear and reliable reasoning path.

\section{Conclusion}
In this work, we propose GraphCheck, a fact-checking method that integrates knowledge graphs to enhance LLM-based fact-checking, particularly for long-form text. GraphCheck addresses the limitations of  LLMs in capturing complex entity relationships, which often result in overlooked factual errors. By leveraging graph neural networks (GNNs) to integrate representations from the generated claim and the source document KGs, our method enables fine-grained fact-checking in a single model call, significantly improving efficiency. Furthermore, the incorporation of graph information enhances the interpretability of the fact-checking process. Experiments on general and medical domain datasets demonstrate that GraphCheck achieves competitive performance.

\section*{Limitations}
\noindent{\textbf{Quality of Knowledge Graphs.}}
Although integrating KGs into the fact-checking process is effective, our method remains limited by the quality of KGs. Currently, there is no reliable method for evaluating the quality of extracted KGs, and the process largely depends on manual judgment. As the dataset grows, it becomes difficult to assess the quality of the extracted KGs. As analyzed in our paper, KG quality directly impacts our method’s performance (errors in the KG may introduce noise or fail to provide sufficient support for fact-checking). 

\noindent{\textbf{Limited Training Data.}}
The lack of high-quality training data is a common limitation in long-text fact-checking. For this reason, and to ensure a fair comparison, we adopted the 14K dataset provided by MiniCheck, which consists of relatively short texts. As shown in the paper,  the performance of our method improves with increased training data, suggesting that there remains significant potential for further improvement if larger or higher-quality long-text datasets are considered.

\section*{Acknowledgments}
Dr. Irene Li is supported by JST ACT-X (Grant JPMJAX24CU) and JSPS KAKENHI (Grant 24K20832). Dr. Qingyu Chen is supported by 1R01LM014604, National Library of Medicine, National Institutes of Health. This work used supercomputers provided by the Research Institute for Information Technology, Kyushu University, through the HPCI System Research Project (Project ID: hp250092). This work is also supported by NVIDIA Academic Grant Program and Google Cloud (Gemma 3 Academic Program).


\bibliography{custom}

\appendix
\onecolumn
\label{sec:appendix}
\section{Benchmark Details}
\label{appendix_benchmark}
\subsection{General Domain Benchmarks}
\noindent{\textbf{AggreFact-XSum, AggreFact-CNN.}} They are subsets of the AGGREFACT benchmark \cite{tang-etal-2023-understanding}, designed for evaluating factual consistency in summarization. These subsets correspond to two widely used summarization datasets: XSum \cite{nallapati2016abstractive} and CNN/DailyMail (CNN/DM) \cite{narayan2018don}, which feature different summarization styles. Both datasets contain relatively long documents, making them well-suited for assessing our method’s effectiveness in handling long-text fact-checking.

\noindent{\textbf{Summeval.}} \cite{10.1162/tacl_a_00373} consists of human evaluations of 16 summarization model outputs based on 100 articles from the CNN/DailyMail dataset. Each summary is rated on a Likert scale from 1 to 5 across four categories: consistency, coherence, fluency, and relevance. In our use of this dataset, we extract each individual claim from the summaries as separate data points. The consistency score is mapped such that a score of 5 is labeled as Support, while scores ranging from 0 to 4 are labeled as Unsupport.

\noindent{\textbf{ExpertQA.}} \cite{malaviya2023expertqa} includes responses from six different systems to expert-curated queries, with sentence-level verification against cited or retrieved documents. In our dataset, a sentence is labeled as Support only if the evidence fully supports it. In contrast, partial and incomplete support is classified as Unsupport.

\subsection{Medical Domain Benchmarks}
\noindent{\textbf{COVID-Fact.}} \cite{saakyan2021covid} is a dataset containing 4,086 claims related to the COVID-19 pandemic. The dataset focuses on automatically detecting true claims and their corresponding source articles, followed by generating counter-claims using automated methods instead of human annotators. 

\noindent{\textbf{PubHealth.}} \cite{kotonya2020explainable} consists of 11,832 claims related to a variety of health topics, including biomedical subjects such as infectious diseases and stem cell research, government healthcare policies like abortion, mental health, and women's health, as well as other public health-related issues. Each claim in the dataset is paired with journalist-crafted, gold-standard explanations that provide judgments to support the corresponding fact-check labels. The dataset is designed for two main tasks: veracity prediction and explanation generation, with claims categorized into four labels: true, false, mixture, and unproven. In our experiments, we use the test set as a benchmark, classifying claims labeled as true as Support, while those labeled as false, mixture, and unproven are classified as Unsupport.

\noindent{\textbf{SCIFact.}} \cite{wadden2020fact} consists of 1,400 expert-written scientific claims, each paired with evidence-containing abstracts annotated with labels and rationales. To construct the dataset, annotators re-formulate naturally occurring claims found in scientific literature—specifically citation sentences—into atomic scientific claims, ensuring clarity and precision. Since its training set is labeled, we use it as a benchmark in our experiments. Furthermore, claims with contradictory evidence or no supporting evidence are classified as Unsupport, while all others are classified as Support.

\subsection{Preprocessing for Benchmark}
\label{sec:benchmark_preprocessing}
\begin{figure*}[t]
    \centering
    \includegraphics[width=1\linewidth]{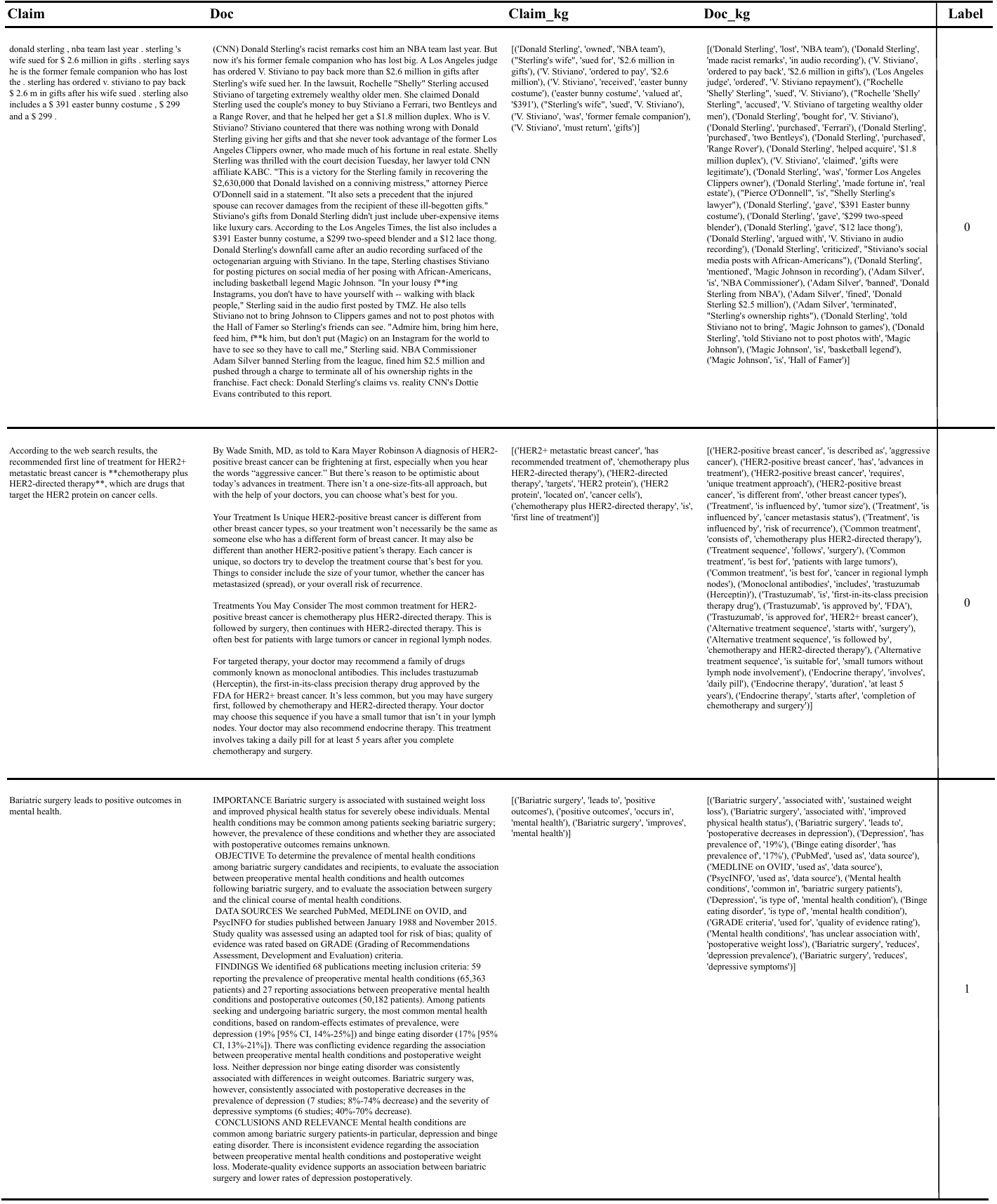}
    \caption{Samples of benchmark data. Each sample consists of a claim, its corresponding document, and the extracted KGs (claim\_kg and doc\_kg), along with the assigned label (Support or Unsupport).}
    \label{fig:benchmark_samples}
\end{figure*}

To extract graph information from the benchmark text data, we utilize LLM to separately extract entity-relation triples from both the claims and the documents. The extraction process follows the prompt shown in \ref{sec:prompts}. After preprocessing, the dataset is structured as \{claim, doc, claim\_kg, doc\_kg, label\}, where claim\_kg and doc\_kg represent the extracted KGs for the claim and document, respectively. Samples of the processed data are illustrated in Figure \ref{fig:benchmark_samples}.

\section{Synthetic Dataset for Training}

\begin{figure*}[t]
    \centering
    \includegraphics[width=1\linewidth]{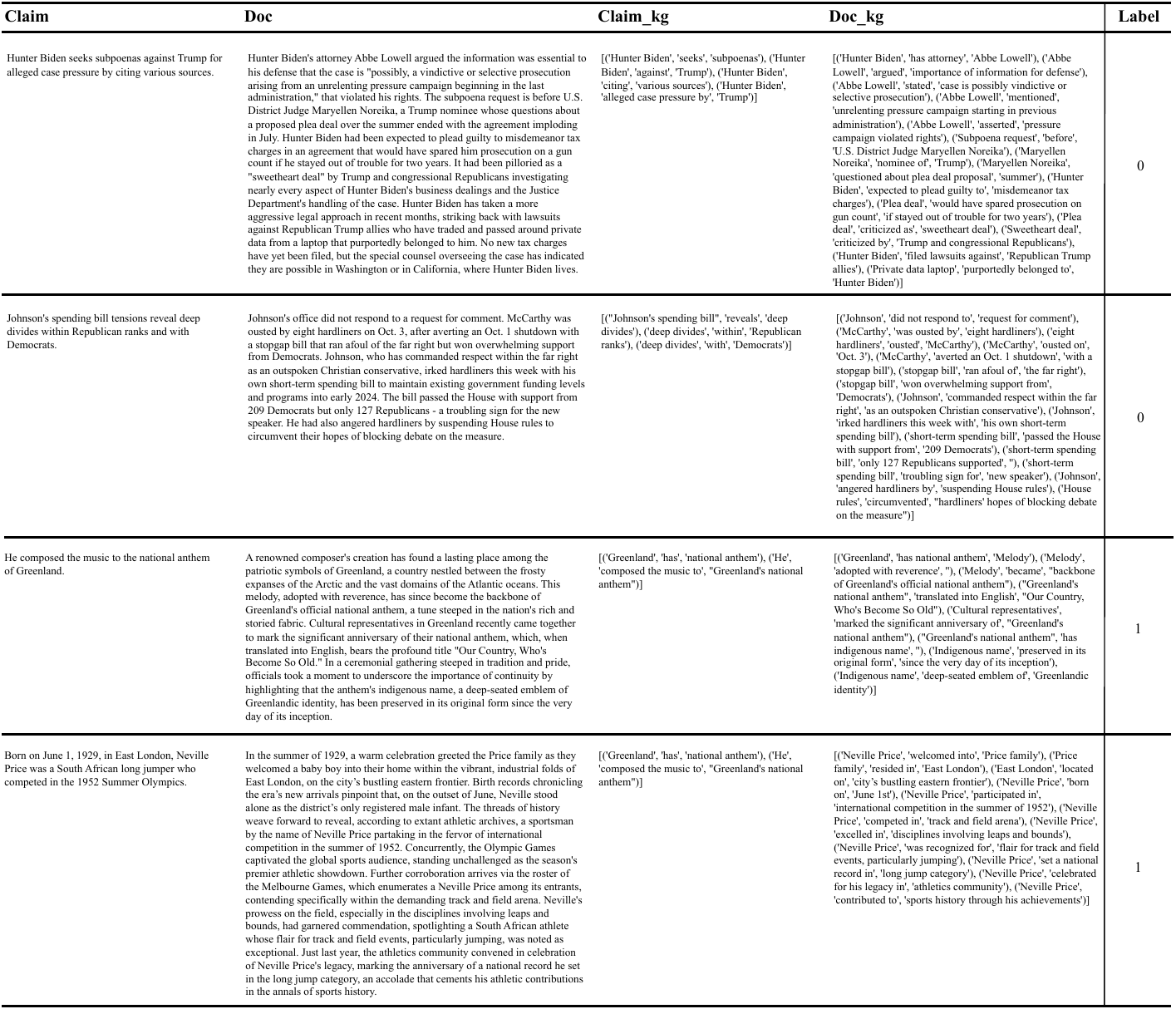}
    \caption{Samples of training data.}
    \label{fig:training_samples}
\end{figure*}

To pre-train an external GNN, we synthesized a structured dataset of 14,000 samples based on the MiniCheck training set. Using a method similar to \ref{sec:benchmark_preprocessing}, we employed the Qwen2.5 7B model to extract KG triples from both the claim and document in each sample, following the prompt in \ref{sec:prompts}. Each sample is structured as \{claim, doc, claim\_kg, doc\_kg, label\}. Examples are shown in Figure \ref{fig:training_samples}.

\section{Analysis of Computational Cost and Time Efficiency}
\begin{table}[t]
    \centering
    \resizebox{0.48\textwidth}{!}{
    \begin{tabular}{lccc}
        \toprule
        Model & \makecell{Avg. Calls\\per Sample} & \makecell{Inference time\\per Sample (secs)} & Cost (\$) \\
        \midrule
        GPT-4       & 1 & 7.1 & 18.6 \\
        OpenAI o1   & 1 & 17.4 & 27.7 \\
        Claude 3.5-Sonnet  & 1  & 8.2 & 7.2 \\
        MiniCheck   & 5 & 0.01 & < 1.0 \\
        ACUEval     & 5 & 5.9 & 8.8 \\
        GraphEval   & 9 & 0.51 & < 1.0 \\
        \midrule
        GraphCheck(Ours)  & 1 & 0.68 & <1.0 \\
        \bottomrule
    \end{tabular}}
    \caption{Comparison of the cost of our method with other specialized fact-checking methods and LLMs.}
    \label{tab:cost}
\end{table}

We compare the computational cost of specialized fact-checking methods and LLMs on the ExpertQA benchmark, selected for its large dataset size and longer text length. For locally deployed methods, we calculate the cost at a rate of \$0.8 per GPU hour, as shown in Table \ref{tab:cost}. The results show that the cost of our method is significantly lower than that of similar LLMs, such as GPT-4, OpenAI O1, and Claude 3.5-Sonnet. Compared to specialized fact-checking methods, our approach shows a substantial efficiency improvement over ACUEval, which also uses the Llama3.3 70B as the base model. Additionally, the cost of our method is comparable to that of Minicheck and GraphEval, which are based on smaller NLI models. Notably, due to the small size of NLI models, their inference speed is fast, allowing Minicheck and GraphEval to maintain low computational costs. However, this also limits their performance and generalization ability. In contrast, our approach remains computationally efficient while achieving superior performance on complex verification tasks. Specifically, our method outperforms Minicheck and GraphEval on the ExpertQA benchmark, demonstrating stronger generalization in handling long-form text scenarios.

\begin{table}[ht]
    \centering
    \resizebox{0.48\textwidth}{!}{
    \begin{tabular}{lcccc}
    \toprule
         \makecell{Graph\_Building\\Method} & XSum & CNN & Summeval & ExpertQA\\
         \midrule
         Edge as Input (used) & 72.9 & 60.3 & 66.2 & 60,3\\
         Edge as Node  & 72.5 & 59.6 & 66.8 & 58.6\\
    \bottomrule
    \end{tabular}}
    \caption{Balanced accuracy comparison of different graph building methods on XSum, CNN, Summeval and ExpertQA benchmarks.}
    \label{tab:edge}
\end{table}

\begin{table}[ht]
    \centering
    \resizebox{0.48\textwidth}{!}{
    \begin{tabular}{ccccc}
    \toprule
         Model & XSum & CNN & Summeval & ExpertQA\\
         \midrule
         Llama3.3 70B        & 60.1 & 53.5 & 57.6 & 54.3\\
         Llama3.3 70B + GAT  & 72.9 & 59.6 & 65.4 & 60.3\\
         Llama3.3 70B + GT   & 64.8 & 62.4 & 67.3 & 59.1\\
    \bottomrule
    \end{tabular}}
    \caption{Balanced accuracy comparison of different GNN architectures on XSum, CNN, Summeval and ExpertQA benchmarks.}
    \label{tab:gnn}
\end{table}

\section{Additional Experiments}
\label{appendix: additional}
\noindent{\textbf{Analyzing the Impact of Different Graph-Building Methods.}}
We explored two different graph-building methods to evaluate the impact of graph building methods on our approach. The first method directly encodes the relation as edge information in the triplet, represented as [entity1, relation, entity2]. The second method treats the relation as a node, represented as [entity1 → relation] and [relation → entity2]. As shown in Table \ref{tab:edge}, the results show that directly encoding edge information leads to slightly better performance compared to treating the relation as a node, although the difference is minimal.

\noindent{\textbf{Impact of Different GNN Architecture.}}
In our study, we explore the effect of different GNN architectures—Graph Attention Network (GAT) \cite{velivckovic2017graph} and Graph Transformer (GT) \cite{yun2019graph}. As shown in Table \ref{tab:gnn}, the experimental results demonstrate that for the XSum dataset, GAT significantly improves performance from the baseline of 60.1 to 72.9\% (+12.8\%), while GT achieves a smaller improvement of 4.7\% (64.8\%). This suggests that XSum relies more on local relationships, where GAT's attention mechanism effectively captures interactions between adjacent nodes. In contrast, GT's global self-attention may introduce noise or lead to over-smoothing, limiting its effectiveness. However, for Summeval and CNN, GT outperforms GAT (Summeval: 67.3\% vs. 65.4\%, CNN: 62.4\% vs. 59.6\%), suggesting that tasks requiring long-range dependencies and global context benefit more from GT's ability to integrate information across the graph structure. For the ExpertQA dataset, both GAT and GT exhibit similar performance.

\begin{table*}[th]
    \centering
    \resizebox{1.0\textwidth}{!}{
    \begin{tabular}{lcccccccccc|c}
    \toprule
         \textbf{Method} & \makecell{\textbf{AggreFact}\\\textbf{-Xsum}} & \makecell{\textbf{AggreFact}\\\textbf{-CNN}} & \textbf{Summeval} & \textbf{ExpertQA} & \textbf{COVID-Fact} & \textbf{SCIFact} & \textbf{PubHealth} &  \makecell{\textbf{Overall} \\Avg. (\%)} \\
         \midrule
         GraphRAG (GPT-4o) \cite{edge2024local}  & 70.0 & 60.4 & 68.2 & 60.7 & 72.7 & 88.2 & 74.2 & 70.6 \\
         \midrule
         GraphCheck-Llama3.3 70B (Ours) & 72.9 & 62.4 & 67.3 & 60.3 & 71.5 & 89.4 & 73.6 & 71.1\\
         GraphCheck-Qwen 72B (Ours) & 72.1 & 66.5 & 71.0 & 57.2 &69.7 & 86.4 &  71.7 & 70.7\\
    \bottomrule
    \end{tabular}}
    \caption{Balanced accuracy of GraphRAG and GraphCheck across all evaluation benchmarks.}
    \label{tab:rag_result}
\end{table*}

\section{GraphRAG Evaluation}

\noindent{\textbf{Implementation Details of GraphRAG.}
\label{graphrag}
To streamline our implementation process, we leveraged the approach from the open-source nano-GraphRAG project\footnote{\url{https://github.com/gusye1234/nano-graphrag}} for our testing phase. In our experiments, we followed the official approach by using GPT-4o and GPT-4o-mini models. GPT-4o was used for planning and generating responses, while GPT-4o-mini was used for summarization tasks. The workflow is divided into two phases: \textbf{Insert} and \textbf{Query}.

In the \textbf{Insert} phase, we insert the extracted document and claim KGs to build the knowledge base.

In the \textbf{Query} phase, GraphRAG retrieves the relevant triples from the knowledge base and invokes GPT-4o for inference to determine whether the claim is supported or not.


We compare GraphRAG and GraphCheck in Table~\ref{tab:rag_result}. 
The results show that GraphRAG, based on GPT-4o, still underperforms our proposed GraphCheck, which is based on Llama3.3 70B and Qwen 72B. Fundamentally, GraphRAG is not the primary focus of our study. Although both involve graph structures, the core methods are different: GraphRAG relies on retrieval-based reasoning, while GraphCheck integrates structured knowledge directly into LLM for fact-checking.
Additionally, GraphRAG is very costly, with expenses significantly exceeding those of direct inference with GPT-4o (reaching \$47.9 in Table 3's scenario), while offering only marginal performance improvements. Since GraphRAG is a general-purpose framework rather than one specifically designed for fact-checking, we believe it may not be the suitable approach for this task.

\section{Extended Discussion on How LLMs Understand Knowledge Graphs}
\label{appendix:discuss}
Apart from fact-checking, using Knowledge Graphs (KGs) to enhance the text understanding ability of LLMs is also a popular research domain. A key challenge in this area is to enable LLMs to effectively understand graph-structured information.
Existing methods~\cite{guo2023gpt4graph, sen-etal-2023-knowledge, baek2023knowledge, tang2024graphgpt} directly convert graph-structured information into natural language as part of the prompt, enabling LLMs to process graph knowledge without requiring additional graph encoders. However, these approach struggles to effectively handle complex KGs, often requiring multiple calls~\cite{sun2023think} to help LLMs fully comprehend the structured information.

\noindent\textbf{GNN-Based Methods.}
To address this limitation, existing methods~\cite{huang2024can, mavromatis2024gnn} use Graph Neural Networks (GNNs) to encode complex KGs, allowing LLMs to focus on the most relevant entities and relationships. Specifically, G-Retriever~\cite{he2024g} uses a GNN to encode the retrieved KG as a graph embedding and converts the graph information into natural language as part of the prompt, feeding both into the LLM for better understanding. GNP~\cite{tian2024graph} encodes KGs into node embedding vectors and extracts the most relevant node embeddings, which are fed into the LLM along with text embeddings.

In this work, to enable efficient fact-checking on long-form texts, we use a pre-trained GNN to separately encode the entity relationships in the extracted claim KG and document KG. The resulting graph embeddings, combined with text prompt embeddings, enhance the LLM’s ability to understand complex logical structures in long texts. On the other hand, directly comparing the extracted document and claim KGs assists the LLM in performing fine-grained fact-checking within a single inference.

\begin{table}[t]
    \centering
    \resizebox{0.3\textwidth}{!}{
    \begin{tabular}{lc}
    \toprule
         Hyperparameter & Value\\
         \midrule
         batch\_size & 8 \\
         num\_epochs & 20 \\
         learning\_rate & 1e{-5} \\
         weight\_decay  & 0.05 \\
         warmup\_epochs  & 2 \\
         early\_stop\_patience & 3\\
         llm\_num\_virtual\_tokens & 4 \\
         max\_txt\_len & 1024\\
         max\_new\_tokens & 5, 8\\
         gnn\_model & gat, gt\\
         gnn\_num\_layers & 2, 3, 4 \\
         gnn\_in\_dim & 1024\\
         gnn\_hidden\_dim & 1024\\
         gnn\_num\_heads & 4\\
         gnn\_dropout & 0.3, 0.4, 0.5\\
    \bottomrule
    \end{tabular}}
    \caption{Hyperparameters.}
    \label{tab:hyperparameters}
\end{table}

\section{Implementation Details}
\label{implementation_details}
For training, we use Llama3.3 70B \footnote{\url{https://huggingface.co/meta-llama/Llama-3.3-70B-Instruct}} and Qwen2.5 72B \footnote{\url{https://huggingface.co/Qwen/Qwen2.5-72B-Instruct}} as the base models, which remain frozen throughout the training process. The trainable external graph encoder is a GNN. We train the models for 20 epochs with early stopping, setting the maximum generation length to 5 and the learning rate to \(1 \times 10^{-5}\). The best model is selected based on performance on the validation set. The experiments are conducted on 4 NVIDIA A100 80GB GPUs for both training and testing. 

\noindent{\textbf{Detail of Hyperparameter.}
We list all the parameters used for both Llama3.3 70B and Qwen2.5 72B models, as shown in table \ref{tab:hyperparameters}. This includes configuration details such as batch size, learning rate, and optimizer settings.

\section{Prompts}
\label{sec:prompts}

\noindent{\textbf{Triplets Extraction.}} 
Figure \ref{fig:prompt1} presents the prompt and an example used for extracting entity-relation triples from a text using an LLM. The example is sourced from the COVID-Fact dataset.

\noindent{\textbf{Fact-Checking.}} 
Figure \ref{fig:prompt2} presents the fact-checking prompt and an example output. Compared to our method (first row), zero-shot LLMs (second row) require additional descriptive instructions to ensure the stability of the generated output format.

\section{Future Work}
Currently, most fact-checking research primarily focuses on English, and there remains a significant gap in multilingual settings~\cite{chataigner2024multilingual, xuan2025mmlu}. To address this limitation, we plan to extend GraphCheck to support multilingual fact-checking in future work.

\begin{figure*}
    \centering
    \includegraphics[width=1\linewidth]{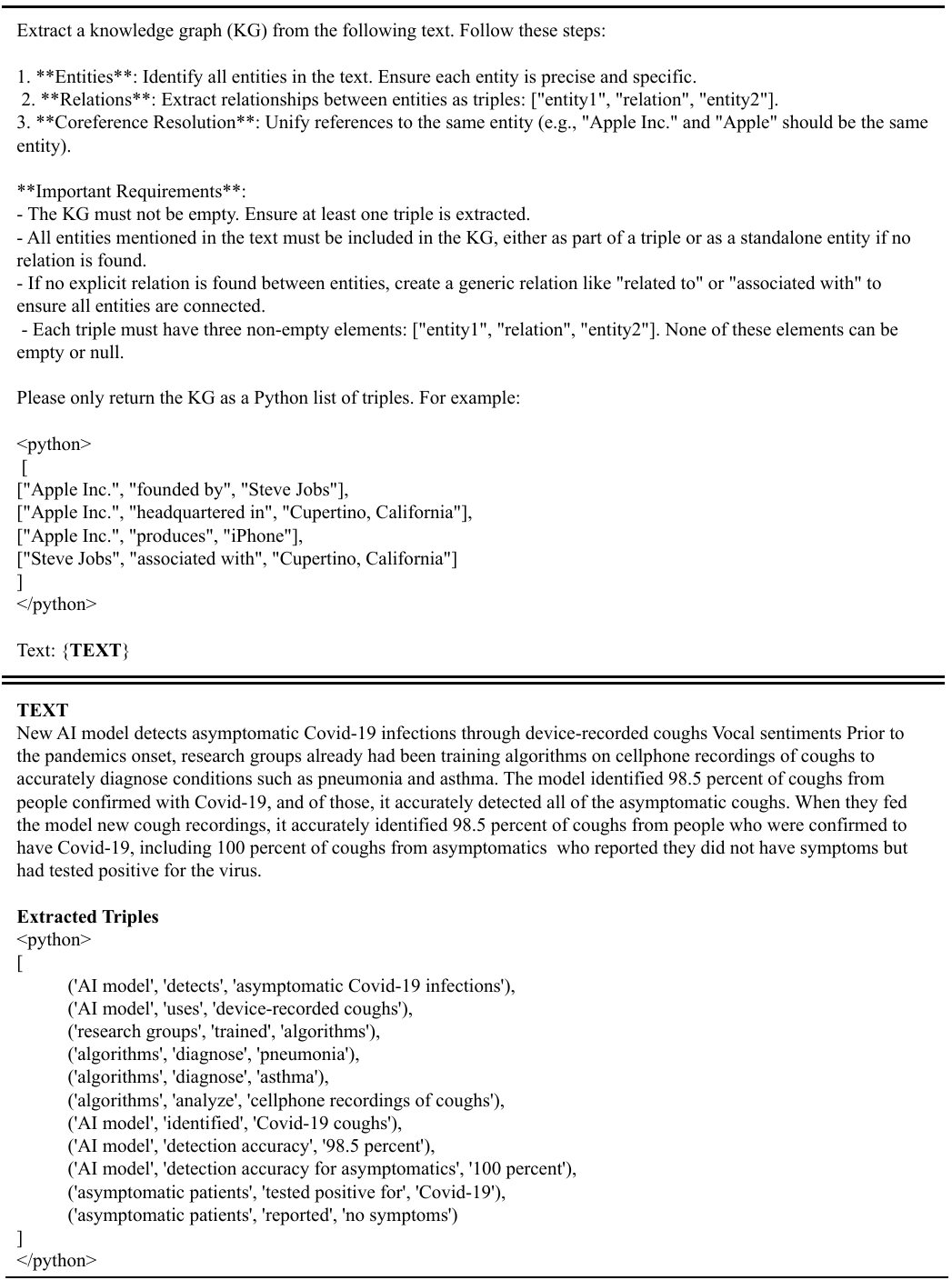}
    \caption{Prompt for Triplets Extraction.}
    \label{fig:prompt1}
\end{figure*}

\begin{figure*}
    \centering
    \includegraphics[width=1\linewidth]{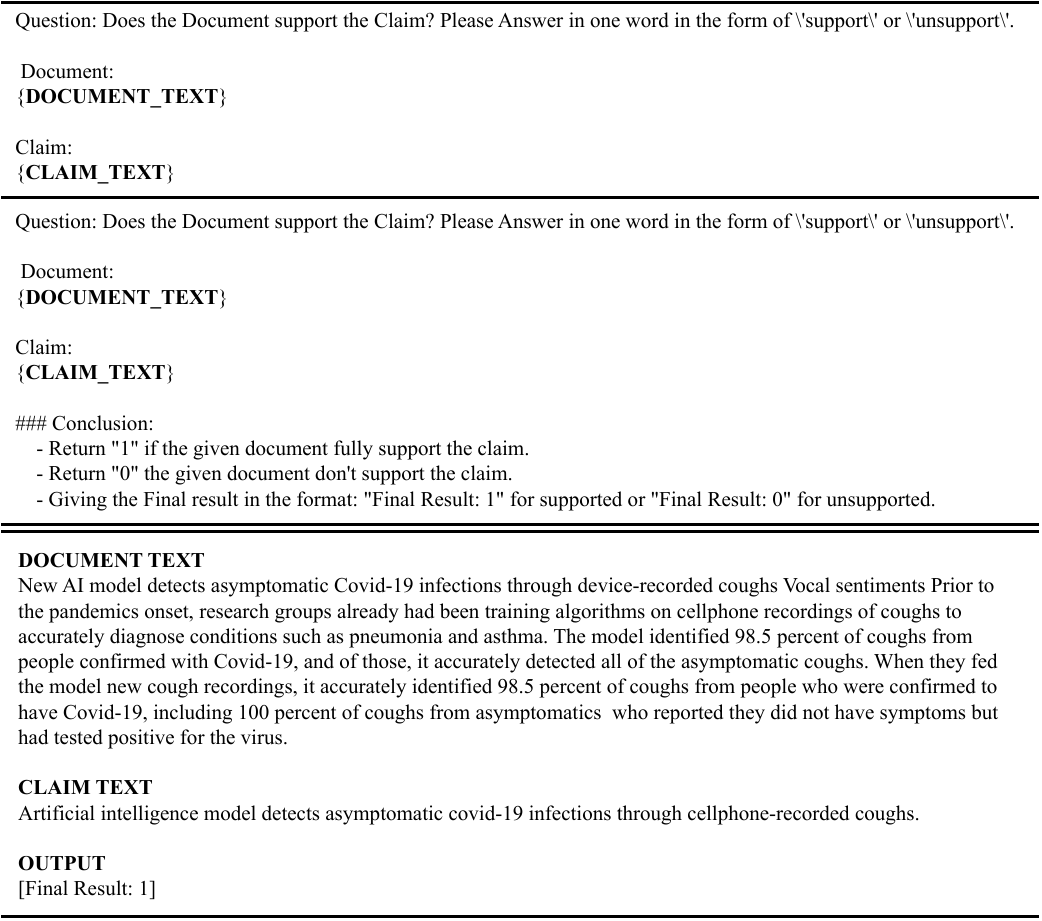}
    \caption{Prompt for Fact-checking.}
    \label{fig:prompt2}
\end{figure*}

\end{document}